\documentclass[letterpaper, 10 pt, conference]{ieeeconf}  

\usepackage{amsmath}
\usepackage{multirow}
\usepackage{graphicx}
\usepackage{caption}
\usepackage{amssymb}
\usepackage{float}
\usepackage[pagebackref=true,breaklinks=true,letterpaper=true,colorlinks,bookmarks=false,urlcolor=red,citecolor=cyan]{hyperref}
\usepackage{cleveref}
\usepackage{xcolor} 
\usepackage{placeins} 
\usepackage{booktabs}
\usepackage{algorithmicx}
\usepackage{algorithm}
\usepackage{algpseudocode}
\usepackage{varwidth}
\usepackage{changepage} 
\usepackage{caption} 
\IEEEoverridecommandlockouts                              

\title{\LARGE \bf
When Should a Robot Think? Resource-Aware Reasoning via Reinforcement Learning for Embodied Robotic Decision-Making
}

\author{Jun Liu$^{1}$, Pu Zhao$^{2}$, Zhenglun Kong$^{3}$, Xuan Shen$^{4}$, Peiyan Dong$^{5}$, Fan Yang$^{6}$, Lin Cui$^{7}$, Hao Tang$^{8}$, \\Geng Yuan$^{9}$, Wei Niu$^{9}$, Wenbin Zhang$^{10}$, Xue Lin$^{2}$, Gaowen Liu$^{12}$,  Dong Huang$^{1}$, Yanzhi Wang$^{2,11}$ 
\thanks{$^{1}$Robotics Institute, Carnegie Mellon University; $^{2}$Northeastern University;  $^{3}$Harvard University; $^{4}$Cornell University; $^{5}$MIT; $^{6}$Fujitsu Research of America;  $^{7}$Tsinghua University; $^{8}$Peking University; $^{9}$University of Georgia; $^{10}$Florida International University;$^{11}$ EmbodyX Inc; $^{12}$Cisco Systems}
}

\begin{document}

\maketitle
\thispagestyle{empty}
\pagestyle{empty}

\begin{abstract}
Embodied robotic systems increasingly adopt \emph{large language model (LLM)-based agents} to support high-level reasoning, planning, and decision-making during interactions with the environment. However, invoking LLM-based reasoning incurs substantial computational latency and resource overhead, which can disrupt action execution and degrade system reliability. Excessive reasoning may delay actions, while insufficient reasoning often leads to incorrect decisions and task failures. A fundamental challenge, therefore, arises: \emph{when should an embodied robot agent think, and when should it act?}
In this work, we propose RARRL (Resource-Aware Reasoning via Reinforcement Learning), a hierarchical framework designed for the resource-aware orchestration of embodied agents. Rather than learning low-level control, RARRL learns an \emph{orchestration policy} that operates at the agent's decision-making layer. This policy enables the embodied agent to adaptively decide whether to invoke reasoning, which reasoning role to employ, and how much computational budget to allocate based on its observations, execution history, and remaining resources.
Extensive experiments—including evaluations using empirical latency profiles from the ALFRED benchmark—demonstrate that RARRL consistently improves task success rates, reduces execution latency, and enhances robustness compared to fixed or heuristic strategies. These results highlight the importance of adaptive reasoning control for reliable and efficient embodied robotic agents.
\end{abstract}

\section{INTRODUCTION}

Embodied robotic systems are increasingly deployed in real-world environments where they must autonomously perceive the environment, reason about task progress, and execute actions under strict time and computational constraints~\cite{siciliano2008springer,thrun2002probabilistic}. %
Recent advances in large language models (LLMs) have significantly enhanced the high-level reasoning and planning capabilities of robotic systems, enabling robots to interpret complex instructions, decompose long-horizon tasks, and adapt to diverse and unstructured scenarios~\cite{brown2020language,wei2022chain, ahn2022can}.
As a result, LLM-based robotic agents have emerged as a promising paradigm for embodied robotic autonomy~\cite{zitkovich2023rt,driess2023palm}.

However, incorporating LLM-based reasoning into embodied robotic systems introduces a critical challenge.
High-level reasoning is computationally expensive and often incurs substantial latency, making it impractical to invoke indiscriminately during task execution~\cite{schulman2017proximal,wang2025large}. %
In real-world robotic deployments, excessive reliance on LLM-based reasoning can delay action execution and disrupt interaction with the environment, while insufficient reasoning frequently leads to incorrect decisions, unsafe behaviors, or task failure~\cite{li2022AdaLoGN,liang2022code}.
This tension between reasoning depth and execution efficiency significantly limits the reliability and responsiveness of embodied robotic agents.

Existing robotic systems typically address this issue using manually designed heuristics or fixed invocation strategies to regulate the use of high-level reasoning modules~\cite{goodrich2008human,kaelbling2011hierarchical}.
Such approaches, however, lack the ability to adapt to varying task complexity, environmental uncertainty, and execution feedback.
Consequently, robots often fail to allocate reasoning resources appropriately across different operational contexts, leading to suboptimal performance and degraded system robustness.
A principled and adaptive mechanism that enables robotic agents to decide \emph{when} and \emph{how} to invoke high-level reasoning during task execution remains largely unexplored.

\begin{figure}[t]
    \centering
    \includegraphics[width=\linewidth]{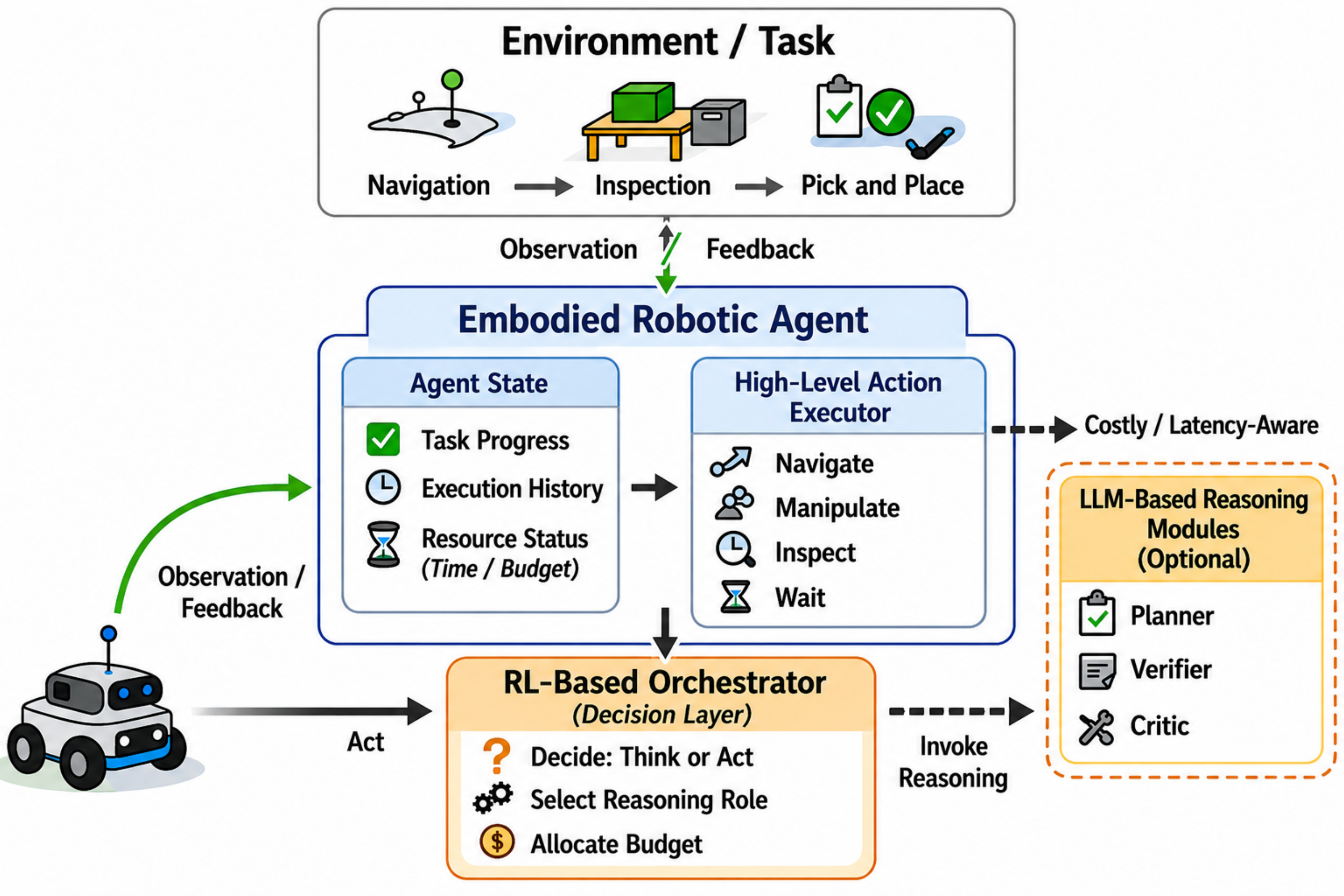}
    \caption{Overview of the proposed single-agent embodied architecture.
     A reinforcement learning (RL) policy operates at the decision-making layer to regulate when the agent should act directly and when to invoke expensive LLM-based reasoning modules under resource constraints.
     The reward signal is used during training to update the orchestration policy based on task outcome and execution latency.
}
    \label{fig:architecture}
\vspace{-0pt} 
\end{figure}

To address these challenges, we propose RARRL (Resource-Aware Reasoning via Reinforcement Learning), an orchestration framework for embodied robotic agents that learns to govern the invocation of LLM-based reasoning modules during environmental interaction. Operating at the decision-making layer of the robotic system, RARRL adaptively manages high-level cognitive processes without modifying low-level perception or motor control.
At each decision step, the learned policy determines whether to invoke high-level reasoning, selects the appropriate reasoning role (e.g., planning or verification), and allocates a suitable computational budget based on the robot's current observations, execution history, and remaining resources~\cite{puterman2014markov,sutton1998reinforcement}.
By explicitly accounting for the cost of reasoning, the proposed approach enables robotic agents to balance reasoning depth and execution efficiency in a data-driven manner.
Figure \ref{fig:architecture} provides an overview of the proposed resource-aware orchestration framework for embodied robotic agents.
The agent interacts with the environment through high-level executable actions, while a reinforcement learning–based orchestrator adaptively decides when and how to invoke costly LLM-based reasoning modules during task execution.

The contributions of this work are summarized as follows:
\begin{itemize}
    \item We identify and formalize a previously underexplored problem in embodied robotic autonomy: \emph{resource-aware decision-making for LLM-based robotic agents}, which concerns how an agent should adaptively decide when and how to invoke high-level reasoning during task execution under limited computational and interaction budgets.
    
    \item We propose a reinforcement learning framework that operates at the agent decision-making layer, learning an orchestration policy over LLM-based reasoning modules without modifying low-level control.
    The learned policy enables the agent to balance reasoning depth and execution efficiency based on its observations, execution history, and remaining resources.
    
    \item We demonstrate through extensive experiments across multiple embodied task scenarios that adaptive reasoning control significantly improves the task success rate, reduces execution latency, and enhances system robustness compared to fixed or heuristic reasoning strategies.
    These results highlight the importance of learning-based reasoning orchestration for reliable and efficient embodied robotic agents.
\end{itemize}

\section{Related Work}

\noindent\textbf{Embodied Decision-Making in Robotics.}
Embodied robotic systems have long been studied as sequential decision-making problems under uncertainty, where agents must perceive, reason, and act through continuous interaction with the environment \cite{thrun2005probabilistic}.
Classical robotics frameworks typically adopt hierarchical architectures that decouple high-level decision-making from low-level perception and control, enabling modularity and interpretability \cite{siciliano2008springer}.
While effective, these approaches often rely on manually designed decision rules or fixed abstractions, limiting their adaptability to complex and dynamic tasks.
Learning-based methods, particularly reinforcement learning (RL), have introduced data-driven alternatives for robotic decision-making and hierarchical control \cite{sutton2018reinforcement}.
However, most RL-based systems focus on learning action-selection policies or task-specific controllers and do not explicitly model the computational cost of high-level reasoning or the dynamic allocation of reasoning resources during execution.

\noindent\textbf{Multimodal Large Language Models for Embodied Reasoning.}
Recent advances in multimodal large language models (MLLMs) have enabled robots to perform high-level semantic reasoning, task planning, and instruction following \cite{wei2022chain,brown2020gpt,li2024embodied,chen2023towards,xiao2026reversible,zhang2026pi,zhou2026comem}.
Prior work has explored grounding language models in robotic action spaces, allowing robots to translate natural language instructions into executable behaviors \cite{ahn2022can,liang2023code}.
More recent systems integrate multimodal language models to support embodied reasoning across perception, language, and action \cite{driess2023palm,zitkovich2023rt}.
Despite their success, most existing LLM-enabled robotic systems invoke high-level reasoning in a static or heuristic manner.
These approaches typically assume that reasoning is always beneficial, without accounting for the substantial computational cost and latency of LLM inference~\cite{chen2025re}.
As a result, excessive reasoning may delay action execution, while insufficient reasoning can lead to task failure, highlighting the need for adaptive reasoning control.

\noindent\textbf{Resource-Aware and Adaptive Control.}
Resource-aware autonomy has been studied in robotics concerning computation, energy, and real-time constraints \cite{mohamed2021energy}.
Existing approaches investigate scheduling, bounded rationality, or task prioritization under limited resources, but often treat reasoning modules as cost-free or rely on static allocation strategies~\cite{eiffert2022resource,xu2022resource}.
FEA~\cite{wu2025efficiency} formalize adaptive reasoning as a control-augmented policy optimization problem balancing task performance with computational cost, distinguishing learned policies from inference-time control mechanisms. %
In the context of LLM-based agents, recent work has emphasized the significant overhead introduced by large-scale reasoning models \cite{xu2025towards}, motivating adaptive mechanisms for regulating reasoning usage.

\noindent\textbf{Reinforcement Learning for High-Level Orchestration.}
Reinforcement learning has been widely applied to robotics for learning control policies and decision-making strategies \cite{schulman2017proximal}.
While hierarchical RL explores multi-level decision structures \cite{kaelbling2011hierarchical}, most prior work focuses on learning behaviors or skills rather than regulating the invocation of reasoning processes.


\begin{figure}[t]
    \centering
    \includegraphics[width=1\linewidth]{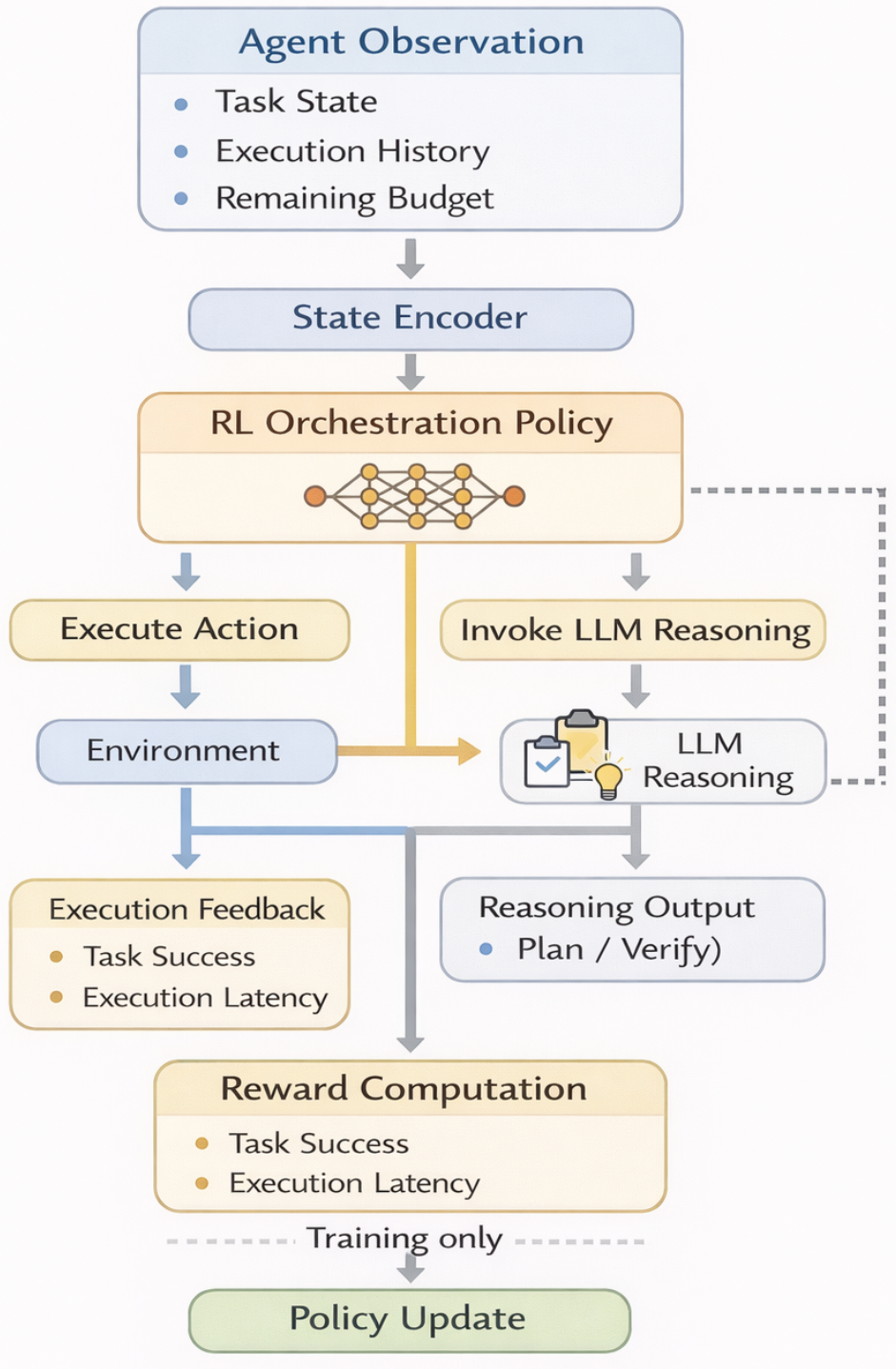}
    \caption{Decision process and training pipeline of the proposed orchestration policy.
The embodied agent observes the current task state, execution history, and remaining computational budget, which are encoded and provided to a reinforcement learning--based policy.
At each decision step, the policy determines whether to execute a high-level action directly or to invoke an LLM-based reasoning module.
Execution feedback, including task outcome and execution latency, is used to compute rewards and update the orchestration policy during training.
}
    \label{fig:pipline}
\vspace{-15pt}
\end{figure}

\section{The Proposed Method}

\subsection{Robot Model}

We model the robot as an embodied agent operating at the execution and decision-making levels.
Rather than simulating low-level dynamics, the robot is represented using a discrete-time abstraction that captures task progress, execution uncertainty, and resource constraints.
This modeling choice reflects the structure of many practical robotic systems, where high-level supervisors issue commands to low-level controllers.

\paragraph{Task Definition}
We consider a representative multi-step delivery task.
Initially, the robot starts from a designated starting location and navigates to a task region, which serves as an intermediate workspace.
At this stage, it is unknown whether deliverable objects are present.
The robot then performs an inspection to identify and confirm a target object.
If a valid object is identified, the robot picks it up and delivers it to a target zone, completing the task.
\paragraph{State}
At time step $t$, the robot state is defined as
\[
s_t = (x_t, h_t),
\]
where $x_t$ denotes the task-related state and $h_t$ encodes recent execution history.

\paragraph{Action Space}
The robot executes a set of high-level actions commonly used in robotic task planning:
\[
a_t \in \{\textsf{Navigate}(r),\ \textsf{Inspect},\ \textsf{Pick},\ \textsf{Deliver}(z),\ \textsf{Wait}\},
\]
where $\textsf{Navigate}(r)$ moves the robot to a specified region $r$,
$\textsf{Inspect}$ inspects the current region to identify task-relevant objects,
$\textsf{Pick}$ attempts to grasp a confirmed object,
and $\textsf{Deliver}(z)$ transports the object to a target zone $z$.
For simplicity, these actions abstract away low-level control details.

\paragraph{State Transition and Latency Modeling}
After executing action $a_t$, the environment returns $(s_{t+1}, \delta_t, o_{t+1}, \textsf{done})$, where $\delta_t$ denotes the wall-clock latency incurred at step $t$.
Latency directly affects the reward via a cost penalty:
$r_t = r_t^{task} - \lambda \delta_t$.
Reasoning actions (\textsc{Think}) incur substantially higher latency than direct execution (\textsc{Act}), reflecting real LLM inference cost.
Thus, frequent reasoning increases cumulative delay and reduces overall return.

Episodes terminate either upon task completion or when a fixed time horizon is reached,
creating an explicit trade-off between task success and responsiveness.


Execution outcomes are stochastic to reflect real-world uncertainty.
For example, navigation may fail with a probability $p_{\text{nav}}$; inspection may produce false positives or negatives; and manipulation actions may fail with a probability $p_{\text{manip}}$.

\paragraph{Reward Function}
The robot receives a sparse task-level reward defined as
\[
r_t =
\begin{cases}
+1, & \text{if the object is successfully delivered},\\
-\lambda \cdot \delta_t, & \text{to penalize execution latency},\\
-\mu, & \text{if an action fails},\\
0, & \text{otherwise},
\end{cases}
\]
where $\lambda$ and $\mu$ control the trade-off between task efficiency and robustness. 

\paragraph{Adaptive Reasoning Control}
At each decision step, the agent may optionally invoke high-level reasoning actions
\[
u_t \in \{\textsf{None},\ \textsf{Plan},\ \textsf{Verify}\},
\]
which incurs additional computational costs but reduces execution uncertainty.
The orchestration policy learns when such reasoning should be invoked to balance task success and execution efficiency.


\subsection{Architecture Overview}
\label{sec:architecture}

Figure~\ref{fig:pipline} shows the overall architecture.
We consider a single embodied agent interacting with an abstract sequential task process through an observation–action interface.
A reinforcement learning (RL) policy governs when to invoke computationally expensive LLM-based reasoning modules, explicitly trading off task performance and resource consumption.
The key idea is to decouple high-level reasoning orchestration from low-level action execution.

\paragraph{RL-Based Orchestration}
At each step, the orchestration policy (the \emph{decider}) selects between two modes:
\textsc{Act}, which directly executes a low-level action, and
\textsc{Think}, which invokes an LLM-based reasoning module (e.g., planning or verification).
Reasoning is modeled as a costly decision and is only invoked when it is expected to improve downstream execution under limited budgets.

\paragraph{Execution and Reasoning Interface}
When the \textsc{Act} is selected, control is passed to a low-level executor.
When the \textsc{Think} is selected, a reasoning role is invoked, and its output is stored as guidance (e.g., high-level instructions or constraints) that condition subsequent execution.
The executor itself remains fixed, and reasoning outputs influence behavior only through this external guidance channel.


\paragraph{Training Signal and Abstraction Note}
During training, the orchestration policy is updated using a reward signal derived from task outcomes and execution costs.
This reward is used solely for learning and is not required at inference time.
We emphasize that the proposed architecture is defined at an abstract decision-making level and does not require real-time interaction with a physical robot or any physics-based simulation.
Additional architecture details, including state aggregation and training signals, are provided in the Appendix~\ref{app:Architecture}.


\subsection{Problem Formulation}
\label{sec:problem}
We formulate adaptive reasoning orchestration as a Markov decision process (MDP),
where an agent must decide \emph{when} to invoke costly high-level reasoning under limited resources.
At each decision step, the policy observes an aggregated state summarizing the current task context,
remaining budget, and execution history, and selects between direct action execution (\textsc{Act})
or invoking a reasoning module (\textsc{Think}) 
The reward encourages task completion while penalizing excessive reasoning costs.
The orchestration policy is trained using reinforcement learning. A full MDP specification, including state, action, transition, and reward definitions,
is provided in the Appendix~\ref{app:problem}.

\subsection{Orchestration Policy Learning}
\label{sec:learning}

We learn the orchestration policy using Proximal Policy Optimization (PPO)~\cite{schulman2017proximal}, a policy gradient method known for its stability in long-horizon decision-making.
Advantages are computed using generalized advantage estimation (GAE)~\cite{schulman2015high}.
Our use of PPO focuses on learning high-level orchestration decisions rather than low-level control.

\paragraph{Policy and Value Parameterization}
The policy $\pi_\theta(a_t \mid s_t)$ is parameterized by a neural network that outputs:
(i) the probability of selecting \textsc{Act} versus \textsc{Think},
(ii) a distribution over reasoning roles $r$ conditioned on \textsc{Think}, and
(iii) a distribution over budget levels $c$.
A value function $V_\phi(s_t)$ is learned jointly to estimate expected returns and serve as a baseline for advantage estimation.
The discrete budget $c \in \{0,1,2\}$ is deterministically mapped to concrete LLM invocation configurations at runtime.
When $c=0$, no LLM call is issued and the executor acts directly.
When $c=1$, only the planner role is invoked with a fixed \texttt{max\_tokens}=256.
When $c=2$, both planner and verifier are sequentially invoked, each with \texttt{max\_tokens}=512.
Thus, $c$ strictly controls (1) the number of LLM forward passes and (2) the maximum generated tokens per environment step.
Across all budget levels, the LLM configuration is fixed (GPT-4o-mini, temperature 0.2, top-$p$=1.0, single sample, no tool calls), so compute differences arise only from role activation and token caps. 
Token limits are enforced via hard API caps, and calls are sequential, so larger $c$ strictly increases latency and token usage, ensuring reproducible and controlled budget scaling.

\paragraph{Training Objective}
We optimize the policy using the clipped PPO objective:
\begin{equation}
\mathcal{L}_{\text{PPO}}(\theta) =
\mathbb{E}_t \left[
\min \left(
\rho_t(\theta) \hat{A}_t,
\mathrm{clip}\big(\rho_t(\theta), 1 - \epsilon, 1 + \epsilon\big) \hat{A}_t
\right)
\right],
\end{equation}
where $\rho_t(\theta) = \pi_\theta(a_t \mid s_t) / \pi_{\theta_{\text{old}}}(a_t \mid s_t)$ and $\hat{A}_t$ denote the advantage estimates.

\paragraph{Training Procedure}
The orchestration policy is trained via \emph{on-policy} interaction with an abstract task model implemented as a purely computational process, rather than a physics-based simulator or physical robot.
At each training iteration, the agent interacts online with this abstract process to collect trajectories using the current policy.
Such processes may be instantiated as programmatic task generators or executable abstractions derived from offline embodied benchmarks.
The LLM-based reasoning modules are treated as black-box components with fixed behavior during training, and only the orchestration policy and value function are updated.
The overall training loop is summarized in Algorithm~\ref{alg:ppo_orchestration}.

\begin{algorithm}[t]
\caption{Orchestration Policy Learning (PPO)}
\label{alg:ppo_orchestration}
\begin{algorithmic}[1]
\For{iteration $=1$ to $N_{\text{iter}}$}
    \State Collect trajectories of length $T$ using $\pi_\theta$
    \State Compute advantages $\hat{A}_t$ using GAE ($\lambda_{\text{GAE}}=0.95$)
    \State Update policy parameters $\theta$ using PPO objective
    \State Update value parameters $\phi$ via mean squared error regression on returns
\EndFor
\end{algorithmic}
\end{algorithm}

By learning a stochastic orchestration policy, the proposed framework adapts reasoning behavior to task context and resource availability without relying on manually designed heuristics.
The separation between orchestration and execution enables compatibility with diverse reasoning backends and abstract interaction processes, making the approach broadly applicable to resource-aware embodied decision-making.

\section{Experiments}

\subsection{Implementation Details}
We implement our framework in Python~3.8 using PyTorch~1.13.0.
All experiments are conducted on an NVIDIA A6000 GPU.
The orchestration policy and value networks are multilayer perceptrons with three hidden layers of 256 units and ReLU activations.

We train the orchestration policy using PPO with standard hyperparameters
($\epsilon=0.2$, learning rate $\eta=3\times10^{-4}$, and $\lambda_{\text{GAE}}=0.95$),
which we found to be robust across tasks.
During training and evaluation, reasoning modules (e.g., LLM-based planners or verifiers) are treated as frozen black boxes.
All results are averaged over 5 random seeds, with standard deviations reported in parentheses. In all experiments, we use $\lambda=1.0$ and $\mu=0.5$.
Additional implementation details and experimental settings are provided in Appendix~\ref{app:implemention}.


\subsection{Robotic Tasks}
\begin{figure}[t]
    \centering
    \includegraphics[width=1\linewidth]{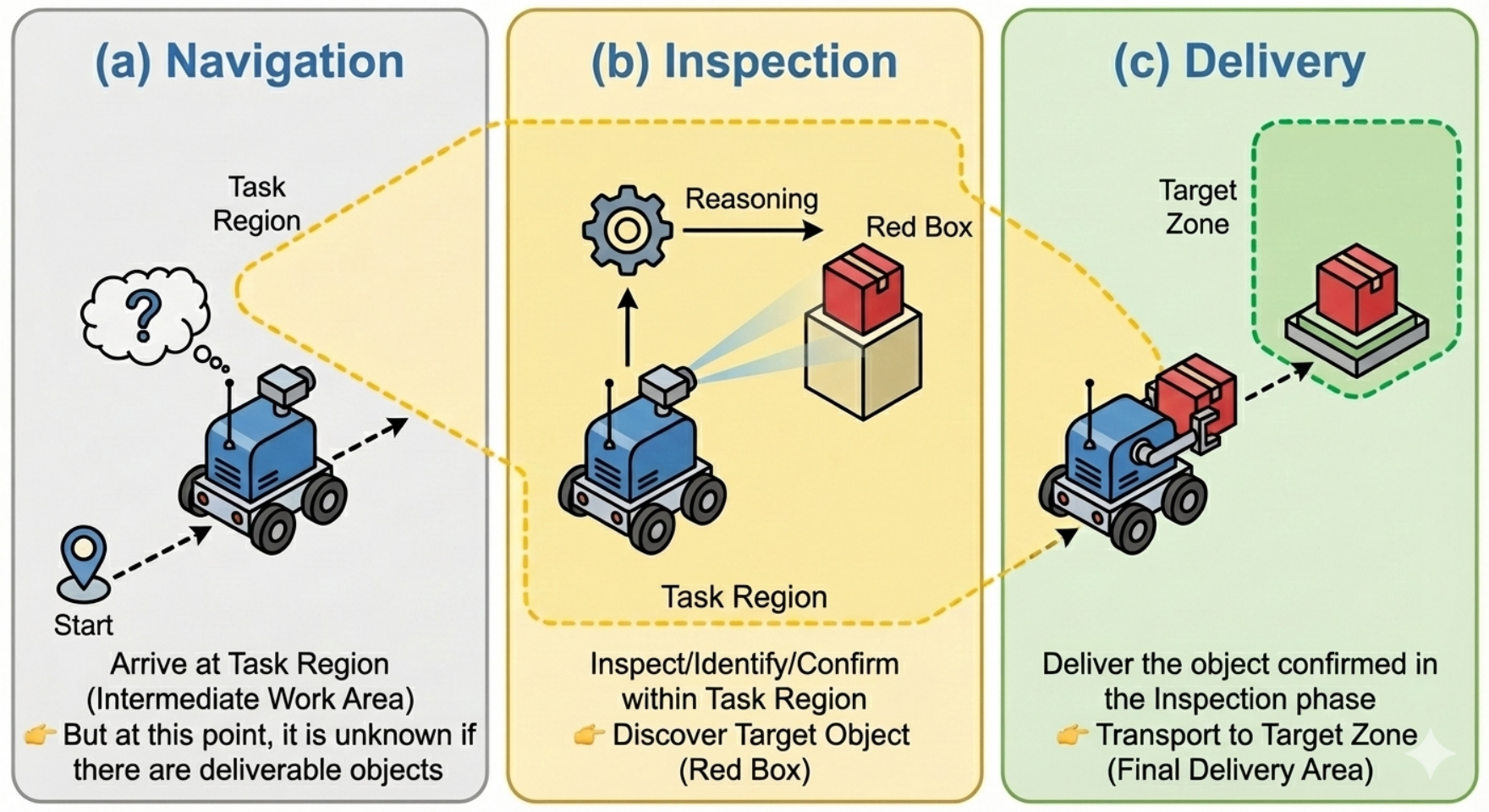}
    \caption{Illustration of a representative multi-step embodied robotic task.
The agent performs navigation, inspection, and delivery stages, during which reasoning demands vary.
Complex stages benefit from high-level reasoning, while routine stages favor direct execution, motivating adaptive reasoning orchestration.}
    \label{fig:task}
    \vspace{-15pt}
\end{figure}

As illustrated in Figure~\ref{fig:task}, we evaluate the proposed approach on embodied robotic tasks composed of multiple execution stages with heterogeneous reasoning demands.
Specifically, we consider:
(1) \emph{Navigation} tasks, where the agent must reach target locations under execution uncertainty;
(2) \emph{Inspection} tasks, which require visiting and verifying objects or regions; and
(3) \emph{Multi-step Delivery} tasks, which sequentially combine navigation, inspection, and object delivery.

These tasks reflect common high-level robotic operations and naturally require adaptive orchestration between direct execution and costly reasoning.

\subsection{ALFRED Runtime Evaluation}
\label{sec:alfred_runtime}

To assess real-world deployment feasibility, we evaluate our orchestration policy in the ALFRED benchmark using the AI2-THOR simulator with real LLM inference in the loop. All baselines use identical LLM configurations and prompt templates.

\begin{table*}[t] 
    \caption{ALFRED runtime evaluation with real LLM inference. Results are reported as mean $\pm$ standard error over 50 episodes per task category. Time is measured in wall-clock seconds. TSR: Task Success Rate (\%), Time: Average wall-clock time per episode (s), LLM-Time: Average LLM inference time per episode (s), Tok: Average token consumption per episode. $^{*}p<0.05$, $^{**}p<0.01$ compared to Constrained PPO (paired t-test).} 
    \label{tab:alfred_runtime} 
    \centering 
    \resizebox{\linewidth}{!}{  
    \setlength{\tabcolsep}{3pt}  
    \begin{tabular}{@{}l*{12}{c}@{}}  
        \toprule 
        & \multicolumn{4}{c}{Navigation} & \multicolumn{4}{c}{Inspection} & \multicolumn{4}{c}{Delivery (Pick--Place)} \\ 
        Method & TSR$\uparrow$ & Time$\downarrow$ & LLM-Time$\downarrow$ & Tok$\downarrow$ & TSR$\uparrow$ & Time$\downarrow$ & LLM-Time$\downarrow$ & Tok$\downarrow$ & TSR$\uparrow$ & Time$\downarrow$ & LLM-Time$\downarrow$ & Tok$\downarrow$ \\ 
        \midrule 
        Full Reasoning~\cite{ramrakhya2025grounding} & 84.0$\pm$1.8 & 42.3$\pm$2.1 & 31.5$\pm$1.9 & 4100$\pm$180 & 78.5$\pm$2.0 & 48.7$\pm$2.4 & 36.8$\pm$2.1 & 4300$\pm$200 & 71.2$\pm$2.3 & 55.1$\pm$2.7 & 40.4$\pm$2.4 & 4500$\pm$220 \\ 
        Heuristic~\cite{thrun2005probabilistic} & 72.4$\pm$2.1 & 27.6$\pm$1.6 & 15.8$\pm$1.2 & 1700$\pm$140 & 66.3$\pm$2.4 & 31.4$\pm$1.8 & 18.1$\pm$1.3 & 1850$\pm$150 & 60.5$\pm$2.6 & 36.9$\pm$2.0 & 21.7$\pm$1.5 & 2100$\pm$170 \\ 
        Constrained PPO & 79.3$\pm$1.9 & 30.8$\pm$1.7 & 18.4$\pm$1.3 & 1950$\pm$150 & 73.1$\pm$2.2 & 34.6$\pm$1.9 & 20.7$\pm$1.4 & 2150$\pm$160 & 65.8$\pm$2.5 & 39.2$\pm$2.1 & 23.9$\pm$1.6 & 2400$\pm$180 \\ 
        Ours (RARRL) & \textbf{82.7$\pm$1.7$^{*}$} & \textbf{25.1$\pm$1.4$^{**}$} & \textbf{12.3$\pm$1.0$^{**}$} & \textbf{980$\pm$90$^{**}$} & \textbf{76.4$\pm$1.9$^{*}$} & \textbf{29.8$\pm$1.6$^{**}$} & \textbf{14.1$\pm$1.1$^{**}$} & \textbf{1100$\pm$100$^{**}$} & \textbf{69.5$\pm$2.1$^{*}$} & \textbf{33.7$\pm$1.8$^{**}$} & \textbf{16.8$\pm$1.3$^{**}$} & \textbf{1350$\pm$120$^{**}$} \\ 
        \bottomrule 
    \end{tabular} 
    }
    \vspace{-6pt} 
\end{table*}
Compared to full reasoning, our orchestration policy reduces LLM inference time by more than 60\% while maintaining comparable task success. 
Relative to heuristic and constrained PPO baselines (Appendix~\ref{app:implemention}), our method achieves consistently higher task success under substantially lower token consumption.
Importantly, the reduction in wall-clock latency directly translates into improved responsiveness in interactive embodied settings.
These results demonstrate that adaptive reasoning orchestration learned in abstract environments can transfer effectively to physics-based simulation with real LLM modules in the loop. We further analyze latency calibration and distribution shift effects in Appendix~\ref{app:calibration_shift}.

\subsection{Main Results on Abstract Tasks}
Table~\ref{tab:main_results} reports results on abstract task scenarios.
The proposed orchestration policy achieves higher task success than heuristic baselines
while substantially reducing reasoning frequency and token usage relative to full reasoning.

Across abstract tasks, our method approaches the success rate of always-on reasoning
while incurring significantly lower computational cost,
demonstrating an improved success--efficiency trade-off in controlled settings.

\begin{table*}[t] 
    \caption{Main results across abstract embodied tasks (mean $\pm$ std over 5 seeds). EL execution latency, RE resource efficiency, RF reasoning frequency, and Tok average LLM token usage per episode. The proposed orchestration policy consistently approaches the success rate of full reasoning while significantly reducing execution latency, reasoning frequency, and token consumption, demonstrating an improved success--efficiency trade-off under limited computational budgets.} 
    \label{tab:main_results} 
    \centering 
    \small 
    \setlength{\tabcolsep}{4pt} 
    \begin{tabular}{@{}lccccc|ccccc@{}} 
        \toprule 
        & \multicolumn{5}{c}{Task Decomposition} & \multicolumn{5}{c}{Structured Navigation} \\ 
        Method & TSR (\%)$\uparrow$ & EL$\downarrow$ & RE$\uparrow$ & RF $\downarrow$& Tok$\downarrow$ & TSR (\%)$\uparrow$ & EL$\downarrow$ & RE$\uparrow$ & RF $\downarrow$& Tok$\downarrow$ \\ 
        \midrule 
        No Reasoning & 45.2$\pm$2.1 & 12.3 & 0.92 & 0.0 & 0 & 52.1$\pm$1.8 & 15.4 & 1.05 & 0.0 & 0 \\ 
        Full Reasoning~\cite{ramrakhya2025grounding} & 85.4$\pm$1.5 & 45.6 & 0.75 & 50.0 & 4200 & 88.3$\pm$1.3 & 48.2 & 0.78 & 50.0 & 4100 \\ 
        Fixed ($k=3$)~\cite{yang2410agentoccam} & 68.7$\pm$1.9 & 28.4 & 0.85 & 16.7 & 1450 & 72.5$\pm$2.0 & 30.1 & 0.88 & 16.7 & 1380 \\ 
        Heuristic~\cite{thrun2005probabilistic} & 72.1$\pm$1.6 & 25.7 & 0.89 & 13.2 & 1120 & 75.4$\pm$1.4 & 27.3 & 0.91 & 13.5 & 1090 \\ 
        Ours(RARRL) & \textbf{82.3$\pm$1.2} & \textbf{18.5} & \textbf{0.95} & \textbf{7.4} & \textbf{620} & \textbf{84.6$\pm$1.1} & \textbf{20.2} & \textbf{0.97} & \textbf{6.9} & \textbf{580} \\ 
        \bottomrule 
    \end{tabular} 
    \vspace{-15pt} 
\end{table*}

\subsubsection{Performance Ceiling Analysis}
\label{sec:ceiling}

\begin{figure}[t]
\centering
\includegraphics[width=0.85\linewidth]{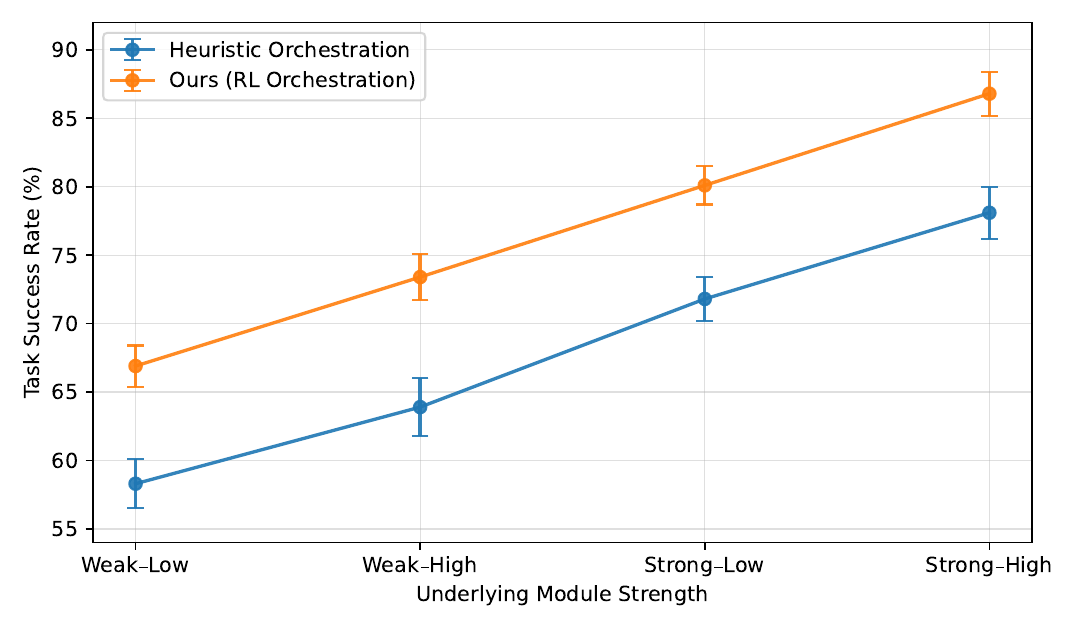}
\caption{Performance ceiling analysis.
Each point shows mean task success over 5 random seeds with standard deviation.
Execution and reasoning strength jointly determine the attainable performance ceiling,
while adaptive orchestration enables closer approach to this ceiling.}
\label{fig:ceiling_curve}
\vspace{-10pt}
\end{figure}

We analyze how orchestration performance is bounded by the strength of the underlying execution and reasoning modules.
Figure~\ref{fig:ceiling_curve} reports task success rates across discrete execution--reasoning configurations.
Increasing execution strength consistently raises the achievable performance ceiling,
indicating that execution fidelity directly constrains maximum task success.
Stronger reasoning further improves performance, particularly when execution is weak,
while under strong execution the marginal gains diminish.
Across all configurations, the proposed orchestration policy consistently outperforms the heuristic baseline,
demonstrating more effective exploitation of available execution and reasoning capacity under resource constraints.

\subsubsection{Robustness and Deployment-Proxy Analysis}
\label{sec:robustness}

\begin{figure}[t]
\centering
\includegraphics[width=0.85\linewidth]{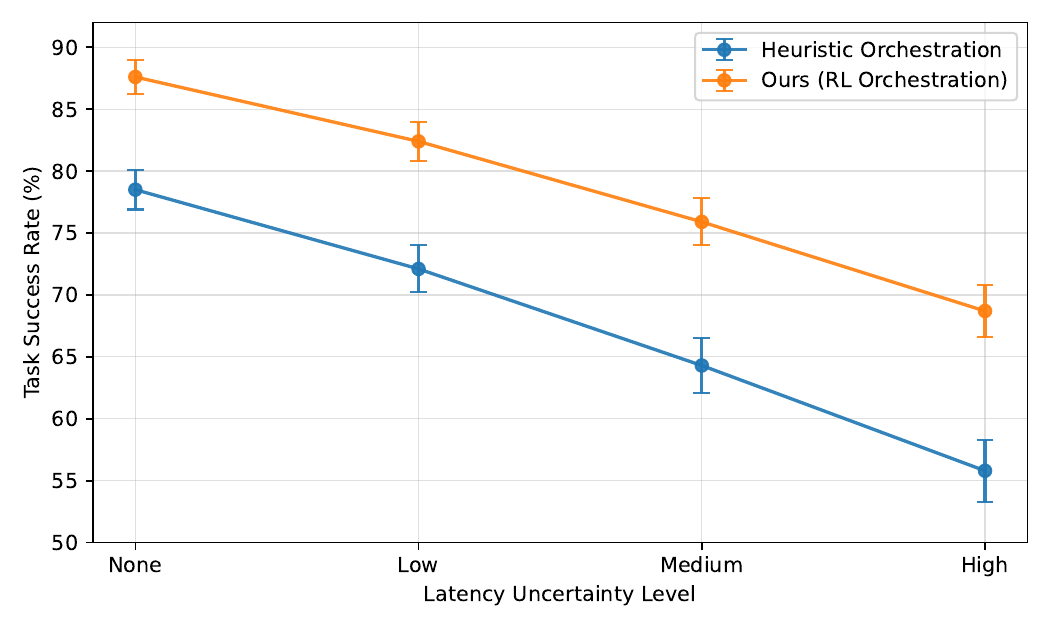}
\caption{Robustness to latency uncertainty.
Task success rate under increasing execution and reasoning latency variability.
Points show mean performance over 5 random seeds, with error bars indicating standard deviation.
The proposed orchestration policy degrades more gracefully than heuristic strategies as latency uncertainty increases.}
\label{fig:robust_latency}
\vspace{-20pt}
\end{figure}

Although trained on abstract interaction processes, real-world deployment involves uncertainty and resource variability.
We evaluate robustness using \emph{deployment-proxy} experiments that inject realistic disturbances into the interaction interface.

\textbf{Latency uncertainty and budget shock.}
We introduce stochastic delays to both action execution and reasoning invocation to simulate execution and computational latency.
As shown in Figure~\ref{fig:robust_latency}, all methods degrade as latency variance increases,
but the proposed policy adapts by reducing unnecessary reasoning and maintains higher task success.

We further evaluate robustness under abrupt resource changes by introducing a \emph{budget shock},
where the remaining computational budget is reduced by a fixed fraction at a random timestep.
Table~\ref{tab:budget_shock} shows that our policy responds by shifting toward action-heavy execution,
preserving substantially higher task success than heuristic baselines that fail to adapt their reasoning rate.


\begin{table}[t]
\caption{Robustness to budget shock.
Task success rate (TSR, \%) and average number of reasoning invocations per episode (RI)
before and after an abrupt reduction of remaining computational budget.
Results are averaged over 5 random seeds.
}
\label{tab:budget_shock}
\centering
\small
\setlength{\tabcolsep}{4pt}
\begin{tabular}{lcc|cc}
\toprule
& \multicolumn{2}{c}{Before Shock} & \multicolumn{2}{c}{After Shock} \\
Method & TSR (\%) $\uparrow$ & RI $\downarrow$ & TSR (\%) $\uparrow$ & RI $\downarrow$ \\
\midrule
Heuristic~\cite{thrun2005probabilistic}
& 78.5 & 14.2
& 61.8 & 15.1 \\
Ours(RARRL)
& \textbf{87.6} & 11.3
& \textbf{74.9} & \textbf{7.6} \\
\bottomrule
\end{tabular}
\vspace{-8pt}  
\end{table}


\begin{table}[t]
\caption{Ablation study on key components of the orchestration policy.
We report task success rate (TSR, \%), average reasoning invocations per episode (RI), and normalized execution cost (Cost; lower is better).
Results are averaged over 5 random seeds.}
\label{tab:ablation}
\centering
\small
\setlength{\tabcolsep}{3pt}
\begin{tabular}{lccc}
\toprule
Variant & TSR (\%) $\uparrow$ & RI $\downarrow$ & Cost $\downarrow$ \\
\midrule
Full (Ours) & \textbf{87.6} & \textbf{11.3} & \textbf{1.00} \\
\midrule
w/o Budget State ($-b_t$) & 80.9 & 15.8 & 1.18 \\
w/o History ($-h_t$) & 78.7 & 13.9 & 1.12 \\
Planner-only (no verifier) & 83.1 & 12.5 & 1.06 \\
Verifier-only (no planner) & 79.4 & 12.1 & 1.05 \\
Fixed Budget (no $c$) & 82.6 & 13.7 & 1.10 \\
\bottomrule
\end{tabular}
\vspace{-20pt} 
\end{table}

\subsection{Ablation Studies}
\label{sec:ablation}

We ablate key design components of the proposed orchestration framework to identify their individual contributions.
Table~\ref{tab:ablation} summarizes results under the same evaluation protocol as the main experiments.
Removing the budget state ($b_t$) yields a notable drop in task success and increases reasoning frequency and execution cost, indicating that explicit resource awareness is essential for preventing over-invocation of expensive reasoning.
Removing execution history ($h_t$) also degrades performance, suggesting that history-aware orchestration decisions help resolve partial observability and recover from past failures.
We further ablate the reasoning-role selection by restricting the policy to a single role.
Planner-only variants outperform verifier-only variants, reflecting the importance of long-horizon guidance, while the full model achieves the best performance by combining complementary planner and verifier signals.
Finally, disabling adaptive budget allocation (fixed $c$) consistently increases cost and reduces success, confirming the benefit of learning fine-grained control over reasoning expenditure.

\section{Discussion}

Our modular formulation highlights key system-level considerations for resource-aware embodied agents. 
First, orchestration performance is bounded by underlying modules: as shown in our ceiling analysis, stronger executors and reasoners raise the attainable success rate, consistent with hierarchical robotic systems where high-level coordination assumes reliable low-level execution.
Second, while we isolate orchestration from sensor noise and actuation delays to enable controlled analysis of reasoning–action trade-offs, extending to physical deployment requires incorporating such uncertainties into state and reward design.
Overall, decoupling orchestration from control yields a modular and deployable design that supports integration with stronger modules and more realistic environments.

Our results empirically validate a broader tradeoff in real-world 
platform deployment: gains in efficiency can conflict with other system 
objectives — a tension notably identified by~\cite{athey2022smiles} in 
the context of online marketplaces, and subsequently observed across 
diverse platform settings including model compression and optimization~\cite{liu2026structured, tan2026harmony, li2025mutual, liu2025toward}, 
efficient multi-agent routing~\cite{liu2025rcr}, autonomous 
systems~\cite{liu2024tsla}, model safety~\cite{zhang2026roots},
online marketplace design~\cite{ma2025balancing}, and platform 
fairness~\cite{ge2022toward, chen2024interpolating}. 
This suggests that adaptive reasoning control is not merely a system 
optimization, but a general design principle for intelligent agents 
operating under resource constraints.

\section{Conclusion}
We presented a reinforcement learning framework that optimizes the use of costly reasoning in embodied agents. Our results confirm that a learned orchestration policy effectively navigates the trade-off between task success and computational latency, maintaining robustness under resource fluctuations. By decoupling high-level decision-making from low-level control, this work provides a scalable foundation for resource-constrained embodied intelligence, moving toward agents that can autonomously determine when to think and when to act.


\clearpage
\section*{APPENDIX}
\subsection{Additional Architecture}
\label{app:Architecture}

\noindent{\textbf{Interaction Interface:}}
The loop consists of an abstract environment with observation and action interfaces.
Observations encode task context, and actions correspond to executable decisions.
We model this interface abstractly, without assuming a physical robot or simulator, enabling instantiation from offline embodied benchmarks (e.g., ALFRED) without hardware interaction.

\noindent\textbf{State Aggregation:}
The agent employs a state aggregation module to fuse multiple information sources into a compact decision state $s_t$. This includes: 
(i) the current observation $o_t$, 
(ii) a resource encoding of remaining computational/interaction budget (e.g., normalized tokens or timesteps), and 
(iii) an execution history of past orchestration decisions and outcomes.
In practice, these are concatenated directly. The history is encoded via a recurrent network or pooling layer, while the resource state uses simple normalization.
In the ALFRED benchmark, the state comprises the instruction embedding, the last action with its success flag, a short history of recent action–outcome pairs, and the normalized remaining budget. No privileged simulator information is used, and no explicit uncertainty estimates are provided; gating decisions rely on observable proxies such as recent failures or action repetitions.
\subsection{{Detailed Problem Formulation}}
\label{app:problem}

We formulate adaptive reasoning control as an MDP, where the policy decides when to invoke reasoning versus act under limited budgets.

\noindent{\textbf{State Space:}}
At step $t$, the state $s_t$ aggregates the current observation $o_t$, resource state $b_t$, and execution history $h_t$.
These components summarize task progress and remaining budget for orchestration.
Observations are encoded as fixed-dimensional embeddings and concatenated with resource and history representations as policy input.

\noindent{\textbf{Action Space:}}
The action space is defined as $\mathcal{A} = \{\textsc{Act}, \textsc{Think}(r, c)\}$,
where $\textsc{Act}$ executes directly and $\textsc{Think}(r, c)$ invokes reasoning with role $r$ and budget $c$.
Roles are selected from a small discrete set (e.g., planner, verifier), and budgets are discretized (e.g., low/medium/high) for stable learning and cost control.

\noindent{\textbf{Effect of THINK on Dynamics:}}
After invoking \textsf{Think}, the failure probability of executing
action $a$ in state $s$ becomes
$P_{fail}^{think}(s,a)=(1-\alpha(s))P_{fail}^{exec}(s,a)$,
where $\alpha(s)\in[0,1]$ denotes the reasoning gain.
In the abstract environment, we model $\alpha(s)$ as a
state-dependent function of measurable execution uncertainty,
$\alpha(s)=\eta\,u(s)$,
where $u(s)\in[0,1]$ is a normalized uncertainty signal
(e.g., recent failure frequency or belief entropy),
and $\eta\in[0,1]$ is a global reasoning effectiveness coefficient.
We treat $\eta$ as an environment parameter and do not learn it during
policy training; the orchestration policy only observes $u(s)$ and the
remaining budget.

\noindent{\textbf{Reward Function:}}
The reward is defined as $r_t = r^{\text{task}}_t - \lambda\, r^{\text{cost}}_t$,
where $r^{\text{task}}_t$ encourages task completion and $r^{\text{cost}}_t$ penalizes reasoning expenditure.
For example, $r^{\text{task}}_t$ may assign a positive reward upon task completion, while $r^{\text{cost}}_t$ is proportional to reasoning time or token usage.
The coefficient $\lambda$ is selected via validation on held-out trajectories, with typical values in the range $[0.1, 1.0]$.

\noindent{\textbf{Objective:}}
The orchestration policy $\pi_\theta(a_t \mid s_t)$ is optimized to maximize the expected discounted return:
$\max_{\theta} \; \mathbb{E}_{\pi_\theta} \left[ \sum_{t=0}^{T} \gamma^t r_t \right],
$
where $\gamma \in (0,1]$ is a discount factor and $T$ is the task horizon.
We adopt soft budget penalties for computational simplicity and stable optimization; extending the formulation to hard constraints via constrained MDPs or Lagrangian methods is left for future work.

\subsection{Additional Implementation Details}
\label{app:implemention}

\noindent{\textbf{Task Scenarios:}}
We consider two abstract embodied task scenarios that capture long-horizon decision-making and interaction structure while remaining independent of physical execution.

\noindent{\textbf{Task Decomposition:}}
A long-horizon instruction-following task where a high-level goal is decomposed into executable sub-actions.
We use a symbolic abstraction derived from embodied benchmarks.

\noindent{\textbf{Structured Navigation:}}
A procedurally generated symbolic navigation task where the agent reaches a target under resource constraints.
Observations encode local structure and goal information, and actions are discrete navigation primitives.
For each task, we generate 1,000 training trajectories and 200 test trajectories using the corresponding abstract task processes.
Each episode has a budget of 50--100 units.
Reasoning costs are sampled from $[5, 20]$ units depending on the selected reasoning role and budget level.
For ALFRED runtime evaluation, we use the official ALFRED validation split and execute tasks in the AI2-THOR physics simulator.
At each reasoning invocation, we query the OpenAI GPT-4o-mini model via API. We fix temperature to 0.2 and cap generation at 512 tokens per call.
Reasoning traces are not exposed to the orchestration policy; the policy only observes the current state features and remaining compute budget. 
Token usage is accumulated per episode and treated as a hard budget constraint during evaluation.
No synthetic latency is injected: all reasoning latency reflects real API response time.
Execution latency corresponds to environment step time in AI2-THOR.

\noindent{\textbf{Training Details.}}
The orchestration policy is trained using PPO with hyperparameters described in Sec.~\ref{sec:learning}.
Episodes last up to $T=50$ steps.
We use $\gamma=0.99$ and $\lambda_{\text{GAE}}=0.95$ across all experiments.

\noindent{\textbf{Baselines:}}
We compare the proposed method against the following baselines:
(1) {No Reasoning}: The agent never invokes high-level reasoning and always executes actions directly.
(2) {Full Reasoning}: The agent invokes reasoning at every decision step~\cite{ramrakhya2025grounding}.
(3) {Fixed Invocation}~\cite{yang2410agentoccam} : The agent invokes reasoning at fixed intervals (every $k$ steps, with $k \in \{3,5\}$ tuned on validation data).
(4) {Heuristic Budget}~\cite{thrun2005probabilistic}: The agent invokes reasoning when a hand-crafted uncertainty signal exceeds a threshold.
All baselines share the same low-level executor and reasoning modules for fair comparison.

\noindent{\textbf{Constrained PPO Baseline:}}
We implement a cost-aware PPO with an explicit episode-level reasoning budget constraint.
Let $C(\tau)$ denote total reasoning cost per episode.
The constraint is:
$
\mathbb{E}[C(\tau)] \le C_{\max},
$
where $C_{\max}$ matches the budget used in our method.
We adopt a Lagrangian formulation:
$
\mathcal{L}(\theta,\lambda)
= \mathbb{E}[R(\tau)]
- \lambda \left(\mathbb{E}[C(\tau)] - C_{\max}\right),
$
with dual update
$\lambda \leftarrow [\lambda + \eta_\lambda (\mathbb{E}[C(\tau)] - C_{\max})]_+$.
We use $\eta_\lambda=0.01$ and keep all other PPO hyperparameters identical to ours.

\noindent{\textbf{Metrics:}}
(1) {Task Success Rate (TSR)}: Fraction of episodes completed successfully.
(2) {Execution Latency (EL)}: Average number of steps per episode, serving as a proxy for execution time (lower is better).
(3) {Resource Efficiency (RE)}: Task success normalized by total budget consumption (higher is better).
(4) {Reasoning Frequency (RF)}: Average number of reasoning invocations per episode.

\subsection{Calibration and Distribution Shift Analysis}
\label{app:calibration_shift}
\noindent{(1) Latency Measurement:}
We measure GPT-4o-mini latency in ALFRED runtime over 300 reasoning calls.
Mean API latency is 0.82s (std 0.27s; 95th percentile 1.34s).
Mean token usage per invocation is 380 tokens.
\noindent{(2) Calibration:}
Before transfer, the abstract reasoning cost model is rescaled to match
the measured mean and variance of runtime latency.
Abstract cost units are linearly mapped to token usage
so that expected per-episode budget matches ALFRED runtime statistics.
No policy fine-tuning is performed after calibration.
\noindent{(3) Transfer Gap:}
After calibration, the abstract-to-runtime performance gap
is below 3\% absolute TSR across task categories.
Token consumption prediction error reduces from 11.3\% to 4.6\%.
\noindent{(4) Shift Robustness:}
Under artificial latency inflation (1.5$\times$),
TSR degrades by less than 2.5\%,
as the policy automatically reduces reasoning frequency.







\begin{thebibliography}{99}


\bibitem{siciliano2008springer}
Siciliano, B., Khatib, O., \& Kr{\"o}ger, T. (2008).
\textit{Springer Handbook of Robotics}.
Springer.

\bibitem{thrun2002probabilistic} S. Thrun, “Probabilistic robotics,” \textit{Communications of the ACM}, vol. 45, no. 3, pp. 52–57, 2002, ACM New York, NY, USA.

\bibitem{brown2020language} T. Brown, B. Mann, N. Ryder, M. Subbiah, et al., “Language models are few-shot learners,” \textit{Advances in neural information processing systems}, vol. 33, pp. 1877–1901, 2020.


\bibitem{wei2022chain} J. Wei, X. Wang, D. Schuurmans,  et al., “Chain-of-thought prompting elicits reasoning in large language models,” \textit{Advances in neural information processing systems}, vol. 35, pp. 24 824–24 837, 2022.

\bibitem{ahn2022can} M. Ahn, A. Brohan, N. Brown, Y. Chebotar, et al., “Do as i can, not as i say: Grounding language in robotic affordances,” \textit{arXiv preprint arXiv:2204.01691}, 2022.

\bibitem{zitkovich2023rt} B. Zitkovich, T. Yu, S. Xu, P. Xu,  et al., “Rt-2: Vision-language-action models transfer web knowledge to robotic control,” in \textit{Conference on Robot Learning}, 2023, pp. 2165–2183, PMLR.

\bibitem{driess2023palm} D. Driess, F. Xia, M. S. M. Sajjadi, C. Lynch, A. Chowdhery, A. Wahid, J. Tompson, Q. Vuong, T. Yu, W. Huang, et al., “Palm-e: An embodied multimodal language model,” 2023.

\bibitem{schulman2017proximal} J. Schulman, F. Wolski, P. Dhariwal, A. Radford, and O. Klimov, “Proximal policy optimization algorithms,” \textit{arXiv preprint arXiv:1707.06347}, 2017.

\bibitem{wang2025large} J. Wang, E. Shi, H. Hu, et al., “Large language models for robotics: Opportunities, challenges, and perspectives,” \textit{Journal of Automation and Intelligence}, vol. 4, no. 1, pp. 52–64, 2025, doi: 10.1016/j.jai.2024.12.003.

\bibitem{sutton2018reinforcement} R. S. Sutton and A. G. Barto, \textit{Reinforcement Learning: An Introduction}, 2nd ed., MIT Press, 2018.

\bibitem{li2022AdaLoGN} X. Li, G. Cheng, Z. Chen, Y. Sun, and Y. Qu, “AdaLoGN: Adaptive Logic Graph Network for Reasoning-Based Machine Reading Comprehension,” in \textit{Proceedings of the 60th Annual Meeting of the Association for Computational Linguistics (Volume 1: Long Papers)}, Dublin, Ireland, 2022, pp. 7147–7161, Association for Computational Linguistics.

\bibitem{liang2022code} J. Liang, W. Huang, F. Xia, P. Xu, K. Hausman, B. Ichter, P. Florence, and A. Zeng, “Code as policies: Language model programs for embodied control,” \textit{arXiv preprint arXiv:2209.07753}, 2022.

\bibitem{puterman2014markov} M. L. Puterman, \textit{Markov decision processes: discrete stochastic dynamic programming}. John Wiley \& Sons, 2014.

\bibitem{sutton1998reinforcement} R. S. Sutton, A. G. Barto, et al., \textit{Reinforcement learning: An introduction}, vol. 1, no. 1. MIT press Cambridge, 1998.

\bibitem{goodrich2008human} M. A. Goodrich, A. C. Schultz, et al., “Human–robot interaction: a survey,” \textit{Foundations and trends® in human–computer interaction}, vol. 1, no. 3, pp. 203–275, 2008, Now Publishers, Inc.

\bibitem{kaelbling2011hierarchical} L. P. Kaelbling and T. Lozano-Pérez, “Hierarchical task and motion planning in the now,” in \textit{2011 IEEE international conference on robotics and automation}, 2011, pp. 1470–1477, IEEE.

\bibitem{li2024embodied}
M. Li, S. Zhao, Q. Wang, K. Wang, Y. Zhou, S. Srivastava, C. Gokmen, T. Lee, E. Li, R. Zhang, W. Liu, P. Liang, L. Fei-Fei, J. Mao, and J. Wu,
``Embodied agent interface: Benchmarking LLMs for embodied decision making,''
\textit{Advances in Neural Information Processing Systems},
vol.~37,
pp.~100428--100534,
2024.
\bibitem{schulman2015high} J. Schulman, P. Moritz, S. Levine, M. Jordan, and P. Abbeel, “High-dimensional continuous control using generalized advantage estimation,” \textit{arXiv preprint arXiv:1506.02438}, 2015.

\bibitem{qian2025xrouter} C. Qian, Z. Liu, S. Kokane, A. Prabhakar, J. Qiu, H. Chen, Z. Liu, et al., “xRouter: Training Cost-Aware LLMs Orchestration System via Reinforcement Learning,” \textit{arXiv preprint arXiv:2510.08439}, 2025.

\bibitem{brown2020gpt} T. B. Brown, B. Mann, N. Ryder, M. Subbiah, J. Kaplan, P. Dhariwal, A. Neelakantan, P. Shyam, G. Sastry, A. Askell, et al., “Language Models are Few-Shot Learners,” in \textit{Advances in Neural Information Processing Systems}, vol. 33, 2020, pp. 1877–1901.

\bibitem{thrun2005probabilistic} S. Thrun, W. Burgard, and D. Fox, \textit{Probabilistic Robotics}. MIT Press, 2005.

\bibitem{liang2023code} Liang, X., et al. "Code as Policies: Language Model Programs for Embodied Control." (2023).

\bibitem{tellex2020foundations} \textsc{Tellex, S.}, et al. ``Foundations of Resource-Aware Robotics.'' \emph{Annual Review of Control, Robotics, and Autonomous Systems}, 2020.

\bibitem{mohamed2021energy} S. A. S. Mohamed, M.-H. Haghbayan, A. Miele, O. Mutlu, and J. Plosila, “Energy-efficient mobile robot control via run-time monitoring of environmental complexity and computing workload,” in \textit{2021 IEEE/RSJ International Conference on Intelligent Robots and Systems (IROS)}, 2021, pp. 7587–7593, IEEE.

\bibitem{xu2025towards} F. Xu, Q. Hao, Z. Zong, J. Wang, Y. Zhang, et al., “Towards large reasoning models: A survey of reinforced reasoning with large language models,” \textit{arXiv preprint arXiv:2501.09686}, 2025.

\bibitem{shi2024opex} H. Shi, Z. Sun, X. Yuan, M.-A. C{\^o}t{\'e}, and B. Liu, “Opex: A component-wise analysis of llm-centric agents in embodied instruction following,” \textit{arXiv preprint arXiv:2403.03017}, 2024.

\bibitem{ramrakhya2025grounding} R. Ramrakhya, M. Chang, X. Puig, et al, “Grounding Multimodal LLMs to Embodied Agents that Ask for Help with Reinforcement Learning,” \textit{arXiv preprint arXiv:2504.00907}, 2025.

\bibitem{yang2410agentoccam} K. Yang, Y. Liu, S. Chaudhary, R. Fakoor, P. Chaudhari, G. Karypis, and H. Rangwala, “Agentoccam: A simple yet strong baseline for llm-based web agents, 2024,” arXiv preprint arXiv:2410.13825.

\bibitem{chen2023towards} L. Chen, Y. Zhang, S. Ren, H. Zhao, et al, “Towards end-to-end embodied decision making via multi-modal large language model: Explorations with gpt4-vision and beyond,” \textit{arXiv preprint arXiv:2310.02071}, 2023.

\bibitem{chen2025re} Y. Chen, H.-J. You, J.-J. Shao, X.-W. Yang, M. Yang,  et al, “$ Re^{} 2$ Agent: Reflection and Re-execution Agent for Embodied Decision Making,” in \textit{NeurIPS 2025 Challenge on Foundation Models for Embodied Agents}, 2025.

\bibitem{eiffert2022resource} S. Eiffert, N. D. Wallace, et al, “Resource and response aware path planning for long-term autonomy of ground robots in agriculture,” \textit{Field Robotics}, vol. 2, pp. 1–33, 2022, FRPS.

\bibitem{xu2022resource} Z. Xu and V. Tzoumas, “Resource-aware distributed submodular maximization: A paradigm for multi-robot decision-making,” in \textit{2022 IEEE 61st Conference on Decision and Control (CDC)}, 2022, pp. 5959–5966, IEEE.

\bibitem{wu2025efficiency} C. Wu, B. Li, M. Gao, Z. Wang, ``From efficiency to adaptivity: A deeper look at adaptive reasoning in large language models,'' \textit{arXiv preprint arXiv:2511.10788}, 2025.

\bibitem{athey2022smiles} S. Athey, D. Karlan, E. Palikot, and Y. Yuan, "Smiles in profiles: Improving fairness and efficiency using estimates of user preferences in online marketplaces," National Bureau of Economic Research, Tech. Rep., 2022.

\bibitem{ge2022toward} Y. Ge, X. Zhao, L. Yu, S. Paul, D. Hu, C.-C. Hsieh, and Y. Zhang, "Toward pareto efficient fairness-utility trade-off in recommendation through reinforcement learning," in \textit{Proceedings of the Fifteenth ACM International Conference on Web Search and Data Mining}, 2022, pp. 316--324.

\bibitem{chen2024interpolating} Q. Chen, J. C. N. Liang, N. Golrezaei, and D. Bouneffouf, "Interpolating item and user fairness in multi-sided recommendations," \textit{Advances in Neural Information Processing Systems}, vol. 37, pp. 50189--50229, 2024.

\bibitem{ma2025balancing} T. Ma, M. S. Bernstein, R. Johari, and N. Garg, "Balancing producer fairness and efficiency via prior-weighted rating system design," in \textit{Proceedings of the International AAAI Conference on Web and Social Media}, vol. 19, Palo Alto, CA: AAAI Press, 2025, pp. 1139--1157.

\bibitem{liu2026structured} J. Liu, Z. Kong, P. Dong, C. Yang, T. Li, Y. Xie, Y. Gong, X. Shen, P. Zhao, H. Tang, et al., "Structured agent distillation for large language model agents," in \textit{Proceedings of the 25th International Conference on Autonomous Agents and Multiagent Systems}, 2026, pp. 3676--3685.

\bibitem{liu2025toward} J. Liu, Z. Kong, P. Zhao, C. Yang, X. Shen, H. Tang, G. Yuan, W. Niu, W. Zhang, X. Lin, et al., "Toward adaptive large language models structured pruning via hybrid-grained weight importance assessment," in \textit{Proceedings of the AAAI Conference on Artificial Intelligence}, vol. 39, no. 18, 2025, pp. 18879--18887.

\bibitem{tan2026harmony} Q. Tan, J. Liu, Z. Zhan, C. Ding, Y. Wang, X. Ma, J. Lee, J. Lu, and G. Yuan, "Harmony in divergence: Towards fast, accurate, and memory-efficient zeroth-order LLM fine-tuning," \textit{Advances in Neural Information Processing Systems}, vol. 38, pp. 175003--175037, 2026.

\bibitem{li2025mutual} S. Li, Q. Tan, Y. Dai, Z. Kong, T. Wang, J. Liu, A. Li, N. Liu, Y. Ding, X. Tang, et al., "Mutual effort for efficiency: A similarity-based token pruning for vision transformers in self-supervised learning," in \textit{Proceedings of the Thirteenth International Conference on Learning Representations}, 2025.

\bibitem{liu2025rcr} J. Liu, Z. Kong, C. Yang, F. Yang, T. Li, P. Dong, J. Nanjekye, H. Tang, G. Yuan, W. Niu, et al., "RCR-router: Efficient role-aware context routing for multi-agent LLM systems with structured memory," \textit{arXiv preprint arXiv:2508.04903}, 2025.

\bibitem{liu2024tsla} J. Liu, Z. Kong, P. Zhao, W. Zeng, H. Tang, X. Shen, C. Yang, W. Zhang, G. Yuan, W. Niu, et al., "TSLA: A task-specific learning adaptation for semantic segmentation on autonomous vehicles platform," \textit{IEEE Transactions on Computer-Aided Design of Integrated Circuits and Systems}, vol. 44, no. 4, pp. 1406--1419, 2024.

\bibitem{zhang2026roots} C. Zhang, Z. Ding, C. Yang, J. Liu, X. Zhai, S. Huang, B. Li, X. Ma, J. Lu, and G. Yuan, "Roots beneath the cut: Uncovering the risk of concept revival in pruning-based unlearning for diffusion models," \textit{arXiv preprint arXiv:2603.06640}, 2026.


\bibitem{xiao2026reversible} C. Xiao, T. Xu, S. Ma, Y. Jiang, H. Gao, and Y. Wu, ``Reversible primitive--composition alignment for continual vision--language learning,'' in \textit{International Conference on Learning Representations}, vol. 2026, pp. 88700--88722, 2026.

\bibitem{zhang2026pi} J. Zhang, C. Zhao, C. Xiao, R. Duan, W. Mo, H. Gao, and W. Wang, ``Pi-cca: Prompt-invariant cca certificates for replay-free continual multimodal learning,'' in \textit{The Fourteenth International Conference on Learning Representations}, 2026.

\bibitem{zhou2026comem} H. Zhou, J. Tang, J. Zhang, Y. Li, C. Xiao, L. Hou, Z. Ke, and J. Yao, ``Comem: Compositional concept-graph memory for vision--language adaptation,'' in \textit{International Conference on Learning Representations}, vol. 2026, pp. 20009--20032, 2026.

















































\end{thebibliography}
\end{document}